# PI-NLF: A Proportional-Integral Approach for Non-negative Latent Factor Analysis

Ye Yuan and Xin Luo, *Senior Member, IEEE*

*Abstract*—A high-dimensional and incomplete (HDI) matrix frequently appears in various big-data-related applications. A non-negative latent factor (NLF) model performs efficient representation learning to an HDI matrix, whose learning process mostly relies on a single latent factor-dependent, non-negative and multiplicative update (SLF-NMU) algorithm. However, an SLF-NMU algorithm updates a latent factor based on the current update increment only without appropriate considerations of past learning information, resulting in slow convergence. Inspired by the prominent success of a proportional-integral (PI) controller in various applications, this paper proposes a Proportional-Integral-incorporated Non-negative Latent Factor (PI-NLF) model with two-fold ideas: a) establishing an Increment Refinement (IR) mechanism via considering the past update increments following the principle of a PI controller; and b) designing an IR-based SLF-NMU (ISN) algorithm to accelerate the convergence rate of a resultant model. Empirical studies on four HDI datasets demonstrate that a PI-NLF model outperforms the state-of-the-art models in both computational efficiency and estimation accuracy for missing data of an HDI matrix.

## I. Introduction

A BIG-DATA-RELATED application commonly involves numerous nodes with the inherent non-negativity interaction relationships, i.e., user-item interactions in a recommender system [1-3], user-user transactions in a blockchain system [4, 5], user-user trusts in a social network system [6, 7], and sensor-sensor communications in IoT [8, 9]. However, it is impossible to obtain their whole interaction relationships due to the exponential growth of involved nodes (e.g., a user touches a tiny subset of items only [10]). In general, a high-dimensional and incomplete (HDI) matrix [11-14] can described such inherent non-negativity interaction relationships, which has only a few known entries (describing the known interactions) while the most others are unknown rather than zeroes (describing the unknown ones). Despite its extreme incompleteness, an HDI matrix contains rich knowledge regarding desired patterns like users' potential favorites [15], item clusters [16], and topological neighbors [17]. Hence, how to efficiently and accurately extract desired knowledge from it for various data analysis tasks [18-20], becomes a highly interesting issue.

Great efforts have been made for addressing this issue, resulting in various data analysis models [21-23]. Among them, a non-negative matrix factorization (NMF) model [21] is ubiquitously adopted to represent a matrix filled with non-negative data. For instance, Li *et al*. [24] propose a weakly-supervised NMF to uncover the relationships between images and tags progressively. Ye *et al*. [25] propose a deep autoencoder-like NMF to map the original network into the community membership space. Rahiche *et al*. [26] propose an orthogonal NMF for accurate multispectral document image segmentation. However, such NMF models must fill the unknown entries of an HDI matrix with zeroes before processing it [27-29], thereby yielding unnecessarily high costs in computation and storage.

To implement convenient and efficient non-negative latent factor analysis on an HDI matrix, Luo *et al*. [30] propose a single latent factor-dependent, non-negative and multiplicative update (SLF-NMU) algorithm, thereby achieving an NLF model. Owing to its dependence on the known part of an HDI matrix only, an NLF model's computational and storage cost is nearly linear with the known entry count of an HDI matrix, which is significantly lower than an NMF model [21-29]. Hence, it has shown great potential on representing an HDI matrix from various applications [30-32].

Although an SLF-NMU algorithm is specifically designed for performing non-negative latent factor analysis on an HDI matrix efficiently, it makes an NLF model suffer slow convergence [30, 33]. This is because it learns a latent factor based on the current update increment only without past information. In industrial applications, a model's convergence rate is vital. Hence, how to further accelerate an NLF model becomes a significant issue.

As described in prior research [34], many widely-adopted optimization algorithms in machine learning [35] share a certain similarity to the classic control mechanisms in automatic control [36]. Both of them aim to reduce the error between the predicted value and desired value. Note that a proportional-integral (PI) controller [37] is the most commonly used control mechanism due to

✧Y. Yuan and X. Luo are with the Chongqing Institute of Green and Intelligent Technology, Chinese Academy of Sciences, Chongqing 400714, China
(e-mail: {yuanye, luoxin21}@cigit.ac.cn).

its simplicity and effectiveness. Specifically, it exploits the current and past errors to make the system stabilize faster [38]. Inspired by this discovery, we have encountered the research question:

***RQ.*** Is it possible to incorporate past update information into SLF-NMU following the principle of a PI controller, thereby accelerating the convergence rate of an NLF mode?

To answer this question, this paper proposes a Proportional-integral-incorporated Non-negative Latent Factor (PI-NLF) model with the following two-fold ideas: a) establishing an Increment Refinement (IR) mechanism with considering the past update increments following the principle of a PI controller; and b) designing an IR-based SLF-NMU (ISN) algorithm to accelerate the convergence rate of a resultant model.

This paper mainly contributes in the following perspective:

a) **A PI-NLF model**. It performs highly efficient and accurate representation learning to HDI data;

Empirical studies on four HDI matrices testify the excellent performance of the proposed PI-NLF model in comparison with state-of-the-art models. Section II states the preliminaries. Section III presents the proposed PI-NLF model. Section IV conducts the empirical studies. Finally, Section V concludes this paper.

## II. PRELIMINARIES

### A. Problem Formulation

***Definition* 1:** Given two large node sets $M$ and $N$, let $R^{|M|\times|N|}$'s each entry $r_{m,n}$ denote the interaction between nodes $m \in M$ and $n \in N$. Let $\Lambda$ and $\Gamma$ denote $R$'s known and unknown node sets, $R$ is an HDI matrix [10-12, 20, 30] if $|\Lambda| \ll |\Gamma|$.

An NLF model tries to build a low-rank approximation to an HDI matrix, which is defined as follows:

***Definition* 2.** Given $R$ and $\Lambda$, an NLF model seeks for a rank-$f$ approximation $\tilde{R}$ to $R$ based on $\Lambda$ with $\tilde{R}=XY^T$ as $X^{|M|\times f}, Y^{|N|\times f} \geq 0$. With the Euclidean distance, such an objective function is formulated by:

$$\varepsilon(X,Y) = \sum_{r_{m,n} \in \Lambda} \left( (r_{m,n} - \tilde{r}_{m,n})^2 + \lambda \left( \|X\|_F^2 + \|Y\|_F^2 \right) \right) = \sum_{r_{m,n} \in \Lambda} \left( \left( r_{m,n} - \sum_{d=1}^f x_{m,d} y_{n,d} \right)^2 + \lambda \left( \sum_{d=1}^f x_{m,d}^2 + \sum_{d=1}^f y_{n,d}^2 \right) \right), \quad (1)$$

$$s.t. \ \forall m \in M, n \in N, d \in \{1,2,...,f\}: x_{m,d} \geq 0, y_{n,d} \geq 0,$$

where $r_{m,n}$, $x_{m,d}$ and $y_{n,d}$ denote the single elements of $R$, $X$ and $Y$, $\tilde{r}_{m,n} = \sum_{d=1}^{f} x_{m,d} y_{n,d}$ denotes the estimate to $r_{m,n} \in \Lambda$ by the NLF model, $\lambda$ denotes the regularization coefficient, and $\|\cdot\|_2$ computes the $L_2$ norm of an enclosed vector, respectively. Note that in (1) we adopt the principle of a data-density-oriented $L_2$ norm-based regularization proposed in [32, 33] to diversify the regularization effects on involved LFs.

### B. An SLF-NMU-based NLF model

As indicated by prior research [30, 33], SLF-NMU is an efficient algorithm to extract non-negative LFs from the known data of an HDI matrix. It firstly applies the additive gradient descent (AGD) to each desired LF, resulting in the following update rule:

$$\underset{X,Y}{\arg\min}\, \varepsilon(X,Y) \overset{AGD}{\Rightarrow} \begin{cases} x_{m,d} \leftarrow x_{m,d} - \eta_{m,d} \sum_{n \in \Lambda(m)} \left( \lambda x_{m,d} + y_{n,d}\tilde{r}_{m,n} - y_{n,d} r_{m,n} \right), \\ y_{n,d} \leftarrow y_{n,d} - \eta_{n,d} \sum_{m \in \Lambda(n)} \left( \lambda y_{n,d} + x_{m,d}\tilde{r}_{m,n} - x_{m,d} r_{m,n} \right); \end{cases} \quad (2)$$

where $\Lambda(m)$ and $\Lambda(n)$ denote the subsets of $\Lambda$ related to $m \in M$ and $n \in N$. $\eta_{m,d}$ and $\eta_{n,d}$ denotes the learning rates corresponding to $x_{m,d}$ and $y_{n,d}$, respectively.

For keeping resultant LFs non-negative, SLF-NMU manipulates $\eta_{m,d}$ and $\eta_{n,d}$ to cancel the negative terms $-\eta_{m,d} \sum_{n \in \Lambda(m)} (\lambda x_{m,d} + y_{n,d}\tilde{r}_{m,n})$ and $-\eta_{n,d} \sum_{m \in \Lambda(n)} (\lambda y_{n,d} + x_{m,d}\tilde{r}_{m,n})$ in (2). Therefore, the learning rules for $x_{m,d}$ and $y_{n,d}$ is given as:

$$\underset{X,Y}{\arg\min}\, \varepsilon(X,Y) \overset{SLF-NMU}{\Rightarrow} \begin{cases} x_{m,d} \leftarrow x_{m,d} \dfrac{\sum_{n \in \Lambda(m)} y_{n,d} r_{m,n}}{\sum_{n \in \Lambda(m)} y_{n,d}\hat{r}_{m,n} + \lambda |\Lambda(m)| x_{m,d}}, \\ y_{n,d} \leftarrow y_{n,d} \dfrac{\sum_{m \in \Lambda(n)} x_{m,d} r_{m,n}}{\sum_{m \in \Lambda(n)} x_{m,d}\hat{r}_{m,n} + \lambda |\Lambda(n)| y_{n,d}}. \end{cases} \quad (3)$$

### C. A PI Controller

Note that a PI controller [37] is the most commonly used feedback control mechanism, which combines the current error and past errors of a feedback control system to calculate a refinement control error, thereby speeding up the stability of the system. In general, it consists of two basic components, i.e., the proportional and integral ones. Its principle is depicted in Fig. 1.

Thus, the general form of a PI controller [37] is given as:

$$\tilde{e}^t = k_p e^t + k_i \int_0^t e^j \, dj, \tag{4}$$

where $e^{(t)}$ denotes the current error at the $t$-th time point, $\tilde{e}^{(t)}$ is the refinement error corresponding to $e^{(t)}$. $k_p$ and $k_i$ denotes the proportional and integral parameter, respectively.

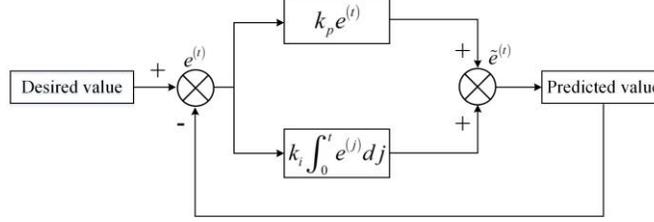

Fig. 1. Flowchart of a PI controller.

Note that we commonly adopt the following discrete form of a PI controller for computational efficiency in practice:

$$\tilde{e}^t = k_p e^t + k_i \sum_{j=0}^{t} e^j. \tag{5}$$

From (5), we can see that the proportional term denotes the current error and integral term denotes the past errors.

## III. A PI-NLF MODEL

### A. An IR Mechanism

According to existing research [36, 37] in automatic control, the goal of a feedback control system is to reduce the error $e(t)$ between the predicted value and desired value by using a controller, thereby achieving a steady state. As shown in Fig. 1, a PI control can reach this goal by calculating a refinement control error $\tilde{e}^{(t)}$ based on the current and past errors.

Note that a machine learning model commonly minimizes its objective function defined on the desired value and the predicted value [35]. Gradient descent (GD) is a widely-adopted optimization algorithm [35] to achieve this goal. Given the decision parameter $\theta$ of the objective function $J$, a GD algorithm updates $\theta$ as follows:

$$\theta_t = \theta_{t-1} - \eta \nabla J(\theta_{t-1}), \tag{6}$$

where $\theta_{t-1}$ and $\theta_t$ denote the states of $\theta$ at the $(t-1)$-th and $t$-th iterations, $\nabla J(\theta_{t-1})$ denotes the gradient at the $(t-1)$-th iteration, $\eta$ denotes the learning rate, respectively.

If the objective function $J$ is not small enough, GD will update the decision parameter $\theta$ based on the gradient $\nabla J(\theta)$. Therefore, it is reasonable to establish the connection between the "gradient" in a machine learning model and the "error" in a PI control as:

$$\nabla \tilde{J}(\theta_{t-1}) = k_p \nabla J(\theta_{t-1}) + k_i \sum_{j=0}^{t-1} \nabla J(\theta_j), \tag{7}$$

where $\nabla \tilde{J}(\theta_{t-1})$ denotes the refinement gradient corresponding to $\nabla J(\theta_{t-1})$.

However, as shown in (3), the SLF-NMU-based learning rules in an NLF model depend on gradients implicitly. Hence, we need to generalize (7) depending on the decision parameter' update increments rather than its gradients. As shown in (6), the update increment by GD is $-\eta \nabla J(\theta_{t-1})$. Actually, for an algorithm implicitly depending on gradients, i.e., SLF-NMU [30, 33], we can also calculate the update increment by the adopted algorithm. Let $\theta'_t$ denote the expected state of the decision parameter on the adopted algorithm at the $t$-th iteration, the update increment can be obtained as follows:

$$\Delta_t = \theta'_t - \theta_{t-1}. \tag{8}$$

By replacing the gradient term in (7) with the update increment in (8), we rewrite (7) into the following form:

$$\tilde{\Delta}_t = k_p \Delta_t + k_i \sum_{j=0}^{t} \Delta_j \Rightarrow \tilde{\Delta}_t = k_p (\theta'_t - \theta_{t-1}) + k_i \sum_{j=0}^{t} (\theta'_j - \theta_{j-1}), \tag{9}$$

where $\tilde{\Delta}_t$ denotes the refinement increment corresponding to $\Delta_t$.

As shown in (9), we establish an IR mechanism following the principle of a PI controller, which considers the past update increments. By substituting such an IR mechanism into (6), a generalized form of (6) can be achieved:

$$\theta_t = \theta_{t-1} + \left( k_p (\theta'_t - \theta_{t-1}) + k_i \sum_{j=0}^{t} (\theta'_j - \theta_{j-1}) \right). \tag{10}$$

### B. An ISN Algorithm

Although SLF-NMU depends on gradients implicitly, its update increment in each iteration can be achieved conveniently. Hence, by combining IR and SLF-NMU, we design the ISN algorithm. Let $X_{t-1}$ and $Y_{t-1}$ denote the status of $X$ and $Y$ after the $(t-1)$-th iteration, and $X'_t$ and $Y'_t$ be the expected states obtained by (3), respectively. Thus, we formulate $X'_t$ and $Y'_t$ as:

$$(X'_t, Y'_t) = \underset{X,Y}{\arg\min}^{SLF\text{-}NMU} \varepsilon(X_{t-1}, Y_{t-1}). \tag{11}$$

Then the update increment by SLF-NMU is given as:

$$\Delta_t = \theta'_t - \theta_{t-1} = \begin{bmatrix} X'_t \\ Y'_t \end{bmatrix} - \begin{bmatrix} X_{t-1} \\ Y_{t-1} \end{bmatrix}. \tag{12}$$

By combining (9) and (12), we achieve that

$$\tilde{\Delta}_t = k_p \left( \begin{bmatrix} X'_t \\ Y'_t \end{bmatrix} - \begin{bmatrix} X_{t-1} \\ Y_{t-1} \end{bmatrix} \right) + k_i \sum_{j=0}^{t} \left( \begin{bmatrix} X'_j \\ Y'_j \end{bmatrix} - \begin{bmatrix} X_{j-1} \\ Y_{j-1} \end{bmatrix} \right). \tag{13}$$

Based on (13), we calculate $\tilde{\Delta}_1$ for the first iteration as:

$$\tilde{\Delta}_1 = k_p \left( \begin{bmatrix} X'_1 \\ Y'_1 \end{bmatrix} - \begin{bmatrix} X_0 \\ Y_0 \end{bmatrix} \right) + k_i \left( \begin{bmatrix} X'_1 \\ Y'_1 \end{bmatrix} - \begin{bmatrix} X_0 \\ Y_0 \end{bmatrix} + \begin{bmatrix} X'_0 \\ Y'_0 \end{bmatrix} - \begin{bmatrix} X_{-1} \\ Y_{-1} \end{bmatrix} \right). \tag{14}$$

where we set $X'_0 = X_{-1}$ and $Y'_0 = Y_{-1}$, and $X_0$ and $Y_0$ denote the initial state of $X$ and $Y$, which is randomly generated non-negative matrices as discussed in [10, 30, 33]. Then $\tilde{\Delta}_1$ can be reformulated as:

$$\tilde{\Delta}_1 = k_p \left( \begin{bmatrix} X'_1 \\ Y'_1 \end{bmatrix} - \begin{bmatrix} X_0 \\ Y_0 \end{bmatrix} \right) + k_i \sum_{j=1}^{1} \left( \begin{bmatrix} X'_j \\ Y'_j \end{bmatrix} - \begin{bmatrix} X_{j-1} \\ Y_{j-1} \end{bmatrix} \right). \tag{15}$$

Afterwards, considering the second iteration, we have:

$$\tilde{\Delta}_2 = k_p \left( \begin{bmatrix} X'_2 \\ Y'_2 \end{bmatrix} - \begin{bmatrix} X_1 \\ Y_1 \end{bmatrix} \right) + k_i \sum_{j=0}^{2} \left( \begin{bmatrix} X'_j \\ Y'_j \end{bmatrix} - \begin{bmatrix} X_{j-1} \\ Y_{j-1} \end{bmatrix} \right). \tag{16}$$

By substituting $X'_0 = X_{-1}$ and $Y'_0 = Y_{-1}$ into (16), we see that

$$\tilde{\Delta}_2 = k_p \left( \begin{bmatrix} X'_2 \\ Y'_2 \end{bmatrix} - \begin{bmatrix} X_1 \\ Y_1 \end{bmatrix} \right) + k_i \sum_{j=1}^{2} \left( \begin{bmatrix} X'_j \\ Y'_j \end{bmatrix} - \begin{bmatrix} X_{j-1} \\ Y_{j-1} \end{bmatrix} \right). \tag{17}$$

Note that situations in the third to $t$-th iterations are the same as that in the prior iterations. Hence, we achieve that

$$\tilde{\Delta}_t = k_p \left( \begin{bmatrix} X'_t \\ Y'_t \end{bmatrix} - \begin{bmatrix} X_{t-1} \\ Y_{t-1} \end{bmatrix} \right) + k_i \sum_{j=1}^{t} \left( \begin{bmatrix} X'_j \\ Y'_j \end{bmatrix} - \begin{bmatrix} X_{j-1} \\ Y_{j-1} \end{bmatrix} \right). \tag{18}$$

By combining (11) and (18), we achieve the training rules of ISN as follows:

$$\begin{bmatrix} X_t \\ Y_t \end{bmatrix} = \begin{bmatrix} X_{t-1} \\ Y_{t-1} \end{bmatrix} + \left( k_p \left( \begin{bmatrix} X'_t \\ Y'_t \end{bmatrix} - \begin{bmatrix} X_{t-1} \\ Y_{t-1} \end{bmatrix} \right) + k_i \sum_{j=1}^{t} \left( \begin{bmatrix} X'_j \\ Y'_j \end{bmatrix} - \begin{bmatrix} X_{j-1} \\ Y_{j-1} \end{bmatrix} \right) \right). \tag{19}$$

Finally, by substituting the learning rule (3) of SLF-NMU into (19), we obtain the following expressions:

$$\begin{cases} x_{m,d_t} = x_{m,d_{t-1}} + \left( k_p \left( \dfrac{x_{m,d_{t-1}} \sum_{n \in \Lambda(m)} y_{n,d_{t-1}} r_{m,n}}{\sum_{n \in \Lambda(m)} y_{n,d_{t-1}} \hat{r}_{m,n} + \lambda |\Lambda(m)| x_{m,d_{t-1}}} - x_{m,d_{t-1}} \right) + k_i \sum_{j=1}^{t} \left( \dfrac{x_{m,d_{j-1}} \sum_{n \in \Lambda(m)} y_{n,d_{j-1}} r_{m,n}}{\sum_{n \in \Lambda(m)} y_{n,d_{j-1}} \hat{r}_{m,n} + \lambda |\Lambda(m)| x_{m,d_{j-1}}} - x_{m,d_{j-1}} \right) \right), \\ y_{n,d_t} = y_{n,d_{t-1}} + \left( k_p \left( \dfrac{y_{n,d_{t-1}} \sum_{m \in \Lambda(n)} x_{m,d_{t-1}} r_{m,n}}{\sum_{m \in \Lambda(n)} x_{m,d_{t-1}} \hat{r}_{m,n} + \lambda |\Lambda(n)| y_{n,d_{t-1}}} - y_{n,d_{t-1}} \right) + k_i \sum_{j=1}^{t} \left( \dfrac{y_{n,d_{j-1}} \sum_{m \in \Lambda(n)} x_{m,d_{j-1}} r_{m,n}}{\sum_{m \in \Lambda(n)} x_{m,d_{j-1}} \hat{r}_{m,n} + \lambda |\Lambda(n)| y_{n,d_{j-1}}} - y_{n,d_{j-1}} \right) \right). \end{cases} \tag{20}$$

Note that when we set $k_p=1$ and $k_i=0$ in (20), the SLF-NMU's learning rules shown in (3) is achieved. Hence, SLF-NMU is a special case of ISN, which only uses the current update increment to update the LFs.

In addition, from (20), we can see that the proportional term denotes the current update increment and integral term denotes the past update increments. Note that the main goal of this study is to verify the effects of incorporating past update increments into the SLF-NMU's training process. Hence, we set $k_p=1$ and (20) can be reformulated into:

$$\begin{cases} x_{m,d_t} = x_{m,d_{t-1}} + \left( \dfrac{x_{m,d_{t-1}} \sum_{n \in \Lambda(m)} y_{n,d_{t-1}} r_{m,n}}{\sum_{n \in \Lambda(m)} y_{n,d_{t-1}} \hat{r}_{m,n} + \lambda |\Lambda(m)| x_{m,d_{t-1}}} - x_{m,d_{t-1}} + k_i \sum_{j=1}^{t} \left( \dfrac{x_{m,d_{j-1}} \sum_{n \in \Lambda(m)} y_{n,d_{j-1}} r_{m,n}}{\sum_{n \in \Lambda(m)} y_{n,d_{j-1}} \hat{r}_{m,n} + \lambda |\Lambda(m)| x_{m,d_{j-1}}} - x_{m,d_{j-1}} \right) \right), \\ y_{n,d_t} = y_{n,d_{t-1}} + \left( \dfrac{y_{n,d_{t-1}} \sum_{m \in \Lambda(n)} x_{m,d_{t-1}} r_{m,n}}{\sum_{m \in \Lambda(n)} x_{m,d_{t-1}} \hat{r}_{m,n} + \lambda |\Lambda(n)| y_{n,d_{t-1}}} - y_{n,d_{t-1}} + k_i \sum_{j=1}^{t} \left( \dfrac{y_{n,d_{j-1}} \sum_{m \in \Lambda(n)} x_{m,d_{j-1}} r_{m,n}}{\sum_{m \in \Lambda(n)} x_{m,d_{j-1}} \hat{r}_{m,n} + \lambda |\Lambda(n)| y_{n,d_{j-1}}} - y_{n,d_{j-1}} \right) \right). \end{cases} \tag{21}$$

For keeping resultant LFs $x_{m,d}^t$ and $y_{n,d}^t$ non-negative, we truncate these two terms to zeroes once they become negative. Based on (21), we achieve the learning rule of ISN for an NLF model.

IV. EXPERIMENTAL RESULTS AND ANALYSIS

A. General Settings

**Evaluation Protocol**. This paper concerns the missing data estimation for an HDI matrix. Hence, we adopt the estimation accuracy as the evaluation protocol [10-12, 20, 30, 32, 33]. Commonly, root mean squared error (RMSE) is commonly adopted to measure a model's estimation accuracy:

$$RMSE = \sqrt{\left(\sum_{r_{m,n} \in \Phi} \left|r_{m,n} - \hat{r}_{m,n}\right|^2\right)/|\Phi|}$$

where |·| calculates the cardinality of an enclosed set, $|\cdot|_{abs}$ denotes the absolute value of an enclosed number, and $\Phi$ denotes testing dataset disjoint with $\Lambda$ and $\Omega$, respectively.

**Datasets.** Four HDI matrices from industrial applications are adopted in our experiments, which are summarized in Table I.

TABLE I. Experimental dataset details.

| No. | Name | Row | Column | Known Entries | Density |
|---|---|---|---|---|---|
| D1 | ML10M [39] | 71,567 | 10,681 | 10,000,054 | 1.31% |
| D2 | Goodbooks [40] | 53,424 | 10,000 | 5,976,480 | 1.12% |
| D3 | Jester [41] | 16,384 | 100 | 1,186,324 | 72.41% |
| D4 | Hetrec-ML [42] | 10,109 | 2,113 | 855,598 | 4.01% |

Note that the known entry set of each HDI matrix is randomly split into ten disjoint and equally-sized subsets, where seven subsets are chosen as the training set, one as the validation set, and the remaining as the testing set. The above process is sequentially repeated five times for five-fold cross-validation. The termination condition is uniform for all involved models, i.e., the iteration threshold is 1000, error threshold is $10^{-5}$, and a model's training process terminates if either threshold is met.

**General Settings**. For achieving the objective results, following general settings are applied to all involved models: a) the LF matrices of each model are initialized with the same randomly generated arrays in each single test on the same dataset to eliminate the initialization bias; b) the dimension of the LF space is set as $f$=20 [10, 30, 33]; c) we fix the regularization coefficient $\lambda$=0.08 for D1-D4; and d) note that as analyzed before, ISN degenerates to the SLF-NMU when $k_p$=1 and $k_i$=0. Therefore, in the next experiments, we mainly verify the effects of $k_i$ with the scale of [0, 0.09] by fixing $k_p$=1.

B. Parameter Sensitivity

In this part, we fix $k_p$=1 to verify the effects of $k_i$. Table II records the iterations and estimation errors corresponding to the iterations. From them, we have the following findings:

TABLE II. PI-NLF's performance with optimal $k_i$ V.S. $k_i$=0.

| Case | | Optimal $k_i$ | | *$k_i$=0 | |
|---|---|---|---|---|---|
| | Value of $k_i$ | Error | Iterations | Error | Iterations |
| D1: RMSE | 0.04 | **0.7988** | **308** | 0.8037 | 588 |
| D2: RMSE | 0.03 | **0.8124** | **230** | 0.8146 | 391 |
| D3: RMSE | 0.04 | **1.0096** | **126** | 1.0114 | 199 |
| D4: RMSE | 0.04 | **0.7695** | **272** | 0.7716 | 485 |

\* The case based on SLF-NMU.

a) **With appropriate $k_i$, PI-NLF's convergence rate is significantly improved by an ISN algorithm.** For instance, as shown in Table II, on D1, PI-NLF consumes 588 to achieve the lowest RMSE with $k_i$=0. In contrast, with $k_i$=0.04, PI-NLF only consumes 308 iterations. The number of iterations decreases at 47.62%. From this view, we can infer that the integral term reflects the past update increments and works like "momentum". It can adjust the updating direction to dampen oscillations, thereby achieving fast convergence. Similar results are found on other datasets.

b) **PI-NLF represents an HDI matrix more precisely than the commonly adopted SLF-NMU-based NLF model.** According to Table II, PI-NLF outperforms the SLF-NMU-based NLF model on all testing cases. For instance, as shown in Table II, on D3, PI-NLF achieves the lowest RMSE 1.0096 and the lowest RMSE of the SLF-NMU-based one is 1.0114. This is because the past update increments maybe help jumping out the local optimum. Similar outcomes also exist on the other datasets.

C. Comparison against State-of-the-art Models

In this part of the experiments, we compare the proposed PI-NLF model with several state-of-the-art models on estimation accuracy and computational efficiency for missing data of an HDI matrix. Table III summarizes the details of all the models. Table IV records their lowest RMSE and converging total time costs. From these results, we have the following important findings:

TABLE III. Details of compared models.

| Model | Description |
|---|---|
| PI-NLF | The proposed model of this study. |
| NLF | An SLF-NMU-based NLF model [30]. |
| WNMF | A weighted NMF model [27]. |
| I-AutoRec | An autoencoder paradigm-based LF model [44]. |
| VAE | A variational autoencoder-based LF model [45]. |
| SIF | A neural interaction functions-based LF model [46]. |
| NueMF | A deep neural network-based LF model [47]. |
| NRT | A recurrent neural network-based LF model [48]. |

a) **PI-NLF's computational efficiency is high when addressing an HDI matrix.** Note that PI-NLF significantly outperforms its peers in terms of computational efficiency on all datasets. For instance, as shown in Table IV, PI-NLF takes 794.6 seconds to achieve the lowest RMSE, which is 53.62% of 1481.8 seconds by NLF, 48.13% of 1650.9 seconds by WNMF, 4.92% of 16166.8 seconds by I-AutoRec, 13.89% of 5722.2 seconds by VAE, 8.97% of 8858.7 seconds by SIF, 0.99% of 79947.6 seconds by NueMF, and 0.83% of 95922.4 seconds by NRT. The same outcomes are also encountered on D2-4.

TABLE IV. Lowest RMSE and their corresponding total time cost (Secs).

| Case | | M1 | M2 | M3 | M4 | M5 | M6 | M7 | M8 |
|---|---|---|---|---|---|---|---|---|---|
| D1: | RMSE | 0.7988±1.3E-4 | 0.8037±2.1E-4 | 0.8893±4.5E-4 | **0.7961±5.5E-4** | 0.8755±2.6E-3 | 0.8852±1.4E-4 | 0.8041±2.2E-4 | 0.8149±4.1E-4 |
| | Time | **794.6±8.2** | 1481.8±75.3 | 1650.9±98.3 | 16166.8±105.2 | 5722.2±89.3 | 8858.7±99.4 | 79947.6±187.4 | 95922.4±278.9 |
| D2: | RMSE | **0.8124±2.4E-4** | 0.8146±2.2E-4 | 0.8835±3.2E-4 | 0.8248±2.3E-4 | 0.9386±4.5E-3 | 0.8758±4.4E-4 | 0.8446±2.5E-3 | 0.8308±6.4E-4 |
| | Time | **484.4±4.3** | 813.3±37.4 | 1423.2±107.6 | 6135.2±28.9 | 632.0±12.8 | 56766.3±325.1 | 30807.4±241.3 | 67176.1±205.8 |
| D3: | RMSE | **1.0096±1.4E-3** | 1.0114±2.1E-3 | 1.0953±2.4E-3 | 1.0161±9.8E-4 | 1.2419±3.4E-4 | 1.1415±1.2E-3 | 1.0344±7.4E-4 | 1.0451±4.9E-4 |
| | Time | **26.7±0.4** | 41.8±1.1 | 53.8±2.5 | 510.5±9.1 | 27.9±0.9 | 7582.3±110.7 | 17502.8±156.9 | 19938.2±86.8 |
| D4: | RMSE | **0.7695±5.6E-4** | 0.7716±4.1E-4 | 0.8390±4.2E-4 | 0.7802±1.3E-3 | 0.8223±7.6E-4 | 0.8124±5.1E-4 | 0.7844±1.4E-4 | 0.7812±4.8E-4 |
| | Time | **72.1±2.3** | 127.6±5.6 | 190.7±4.2 | 500.8±15.2 | 632.2±32.4 | 4594.2±99.6 | 5336.5±200.3 | 6217.6±155.2 |

b) **PI-NLF accurately recovers missing data of an HDI matrix.** On D2-4, PI-NLF has the highest estimation accuracy. For instance, on D2, PI-NLF's RMSE is 0.27 % lower than 0.8146 by NLF, 8.75% lower than 0.8835 by WNMF, 1.53% lower than 0.8248 by I-AutoRec, 15.53% lower than 0.9386 by VAE, 7.80% lower than 0.8758 by SIF, 3.96% lower than 0.8446 by NueMF, and 2.26% lower than 0.8308 by NRT. In addition, PI-NLF is only outperformed by I-AutoRec on D1. Hence, PI-NLF's ability to estimate an HDI matrix's missing data is impressive.

## V. CONCLUSIONS

This paper proposes a PI-NLF model. It incorporates the past update increments following the principle of a PI controller for high convergence rate and representation learning ability to HDI data. In the future, we plan to address the following issues: a) it is necessary to implement self-adaptation of hyper-parameters via advanced evolutionary computation algorithms [43, 50] for excellent practicability; b) it is interesting to investigate other control mechanisms [51] and associate them with more optimization algorithms [35] in machine learning; and c) theoretical analysis on PI-NLF's convergence ability is needed.